\documentclass[sigconf]{acmart}
\AtBeginDocument{%
  \providecommand\BibTeX{{%
    \normalfont B\kern-0.5em{\scshape i\kern-0.25em b}\kern-0.8em\TeX}}}

\usepackage{amsmath}
\usepackage{graphicx}
\usepackage{CJKutf8}
\usepackage{multirow}
\usepackage{adjustbox}

\begin{document}

\title{MDPE: A Multimodal Deception Dataset with Personality and Emotional Characteristics}

\author{Cong Cai}
\affiliation{
  \institution{Beijing Institute of Technology}
  \city{Beijing}
  \country{China}
  }
\email{caicong@bit.edu.cn}  

\author{Shan Liang}
\affiliation{
  \institution{Xi'an Jiaotong Liverpool University}
  \city{Suzhou}
  \country{China}
  }
\author{Xuefei Liu}
\affiliation{
  \institution{Institute of Automation, Chinese Academy of Sciences (CAS)}
  \city{Beijing}
  \country{China}
  }
\author{Kang Zhu}
\affiliation{
  \institution{Anhui University}
  \city{Hefei}
  \country{China}
  } 
\author{Zhengqi Wen}
\affiliation{
  \institution{Beijing National Research Center for Information Science and Technology, Tsinghua University}
  \city{Beijing}
  \country{China}
  }  
  
\author{Jianhua Tao}
\affiliation{
  \institution{Department of Automation, Tsinghua University}
  \city{Beijing}
  \country{China}
  }

\author{Heng Xie}
\affiliation{
  \institution{Beijing Institute of Technology}
  \city{Beijing}
  \country{China}
  }
\author{Jizhou Cui}
\affiliation{
  \institution{ShanghaiTech University}
  \city{Shanghai}
  \country{China}
  }
\author{Yiming Ma}
\affiliation{
  \institution{University of Chinese Academy of Sciences}
  \city{Beijing}
  \country{China}
  } 
\author{Zhenhua Cheng}
\affiliation{
  \institution{University of Chinese Academy of Sciences}
  \city{Beijing}
  \country{China}
  }  
\author{Hanzhe Xu}
\affiliation{
  \institution{Tianjin Normal University}
  \city{Tianjin}
  \country{China}
  }

\author{Ruibo Fu}
\affiliation{
  \institution{Institute of Automation, CAS}
  \city{Beijing}
  \country{China}
  }
\author{Bin Liu}
\affiliation{
  \institution{Institute of Automation, CAS}
  \city{Beijing}
  \country{China}
  }
\author{Yongwei Li}
\affiliation{
  \institution{Institute of Psychology, CAS}
  \city{Beijing}
  \country{China}
  }

\begin{abstract}
Deception detection has garnered increasing attention in recent years due to the significant growth of digital media and heightened ethical and security concerns. It has been extensively studied using multimodal methods, including video, audio, and text. In addition, individual differences in deception production and detection are believed to play a crucial role. Although some studies have utilized individual information such as personality traits to enhance the performance of deception detection, current systems remain limited, partly due to a lack of sufficient datasets for evaluating performance. To address this issue, we introduce a multimodal deception dataset MDPE \footnote{The dataset is available at https://github.com/cai-cong/MDPE.}. Besides deception features, this dataset also includes individual differences information in personality and emotional expression characteristics. It can explore the impact of individual differences on deception behavior. It comprises over 104 hours of deception and emotional videos from 193 subjects. Furthermore, we conducted numerous experiments to provide valuable insights for future deception detection research. MDPE not only supports deception detection, but also provides conditions for tasks such as personality recognition and emotion recognition,  and can even study the relationships between them. We believe that MDPE will become a valuable resource for promoting research in the field of affective computing.
\end{abstract}





\maketitle

\section{Introduction}

Generally, deception refers to the act of misleading, tricking, or deceiving others \cite{depaulo2003cues}. It involves hiding the truth or presenting false information to create an impression that is not accurate. Deception can take many forms, including both verbal and nonverbal information \cite{burgoon2021nonverbal}. And it also occurs in various contexts, such as interpersonal relationships, business, politics, and entertainment. Deception is often considered unethical and can have serious consequences for trust and relationships. 

As deception has expanded to other fields such as social media, interviews, online transactions, the need arises for a reliable and efficient system to aid the task of detecting deceptive behavior. Many machine learning approaches have been proposed in order to improve the reliability of deception detection systems \cite{granhag2008new}. In particular, physiological, psychological, visual, linguistic, acoustic, and thermal modalities have been analyzed in order to detect discriminative features and clues to identify deceptive behavior \cite{feng2012syntactic,hirschberg2005distinguishing,newman2003lying,rajoub2014thermal}. Video-based deception detection is a current priority in deception research, because behavioral cues can be extracted from videos in a cheaper, faster, and non-invasive manner \cite{burzo2018multimodal}, which is preferable to invasive approaches that extract clues through devices attached to human bodies (e.g., polygraphs). Visual clues of deception include facial emotions, expression intensity, hands and body movements, and microexpressions. These features were shown to be capable of discriminating between deceptive and truthful behavior \cite{ekman2009telling,owayjan2012design}. Recently, multimodal analysis has gained a lot of attention due to their superior performance compared to the use of unimodal modalities.  In the deception detection field, several multimodal approaches \cite{perez2015verbal,krishnamurthy2018deep,csen2020multimodal,mathur2020introducing} have been suggested to improve deception detection by integrating features from different modalities. This integration created a more reliable system that is not susceptible to factors affecting sole modalities and polygraph tests.

Substantial empirical evidence indicates significant individual differences in both deception production and detection capabilities \cite{levitan2015cross,majumder2017deep,ren2021sentiment}. These differences encompass cognitive processing, personality traits, psychological characteristics, and emotional expressivity. Empirical studies confirm that personality factors and emotional cues critically influence subjects' capacity to deceive and detect deception \cite{levitan2015cross,gaspar2013emotion}. Emotion—a fundamental dimension of human communication—interacts with cognition to guide social behavior across both interpersonal and human-computer interactions \cite{gordon2016affective,marchi2015voice}. This relationship is particularly relevant to deception, as deceptive acts can elicit distinctive emotional states that manifest as behavioral clues \cite{ekman2009telling,vrij2008detecting}. However, leveraging emotional features to improve deception detection accuracy remains challenging \cite{hartwig2014lie}. A primary complicating factor is that emotional expression itself constitutes a core component of deception, making it difficult to discern whether a deceiver’s displayed emotions are genuine or strategically fabricated.

To address this issue, we propose a multimodal deception dataset MDPE. It not only collects subjects' deception information, but also personality information and emotional information. Each subject was required  to conduct another emotional experiment in addition to engaging in deception, in order to obtain their true emotional expression. Although our research was conducted in the laboratory to provide clear and comparable conversations, we provided subjects with effective monetary incentives to detect and generate effective deceptive behavior \cite{levitan2015cross}. To our knowledge, this is the largest multimodal deception dataset in the released dataset and the only deception detection dataset with personality and emotional characteristics.

To sum up, our contributions are threefold:

\begin{itemize}

\item  We propose a novel multimodal deception dataset MDPE with personality and emotional characteristics, composed of facial video, and audio recordings and transcript. And an easily replicable experimental protocol has also been provided to researchers.
\item  We provide a benchmark for deception detection from multimodal signals, and discussed the impact of personality traits and emotional cues on deception detection.
\item  We offer new possibilities to facilitate further affective computing research, encourage the development of new methods that utilize individual differences for deception detection, as well as for tasks such as personality recognition and emotion recognition.

\end{itemize}

\begin{table}[h]
\begin{center}
\caption{Comparison of the subject count and length for several databases for deception detection }
\begin{tabular}{ccc}
\hline
dataset     & Subjeet Count & Length(Minutes) \\ \hline
Multimodal  & 30            & -               \\
Real Trials & 56            & 56              \\
Box-of-Lics & 26            & 144             \\
Bag-of-Lies & 35            & \textless{}241  \\
DDPM        & 70            & 776             \\ \hline
MDPE        & 193           & 6209            \\ \hline
\end{tabular}
\end{center}
\vspace{-0.34cm}
\end{table}

\section{Related Work}

\textbf{Deception Dataset} \  P{\'e}rez-Rosas et al. \cite{perez2015verbal} introduced a new multi-modal deception dataset Real-life Trial  having real-life videos of courtroom trials. They demonstrated the use of features from different modalities and the importance of each modality in detecting deception. The Box-of-Lies dataset \cite{soldner2019box} was released with video and audio from a game show, and presents preliminary findings using linguistic, dialog, and visual features. Multiple modalities have been introduced in the hope of enabling more robust detection. P{\'e}rez-Rosas et al. \cite{perez2014multimodal} introduced a dataset for deception including video and thermal imaging, as well as physiological and audio recordings. Gupta et al. \cite{gupta2019bag} proposed Bag-of-Lies, a multimodal dataset with gaze data for detecting deception in casual settings. Speth Jeremy et al.\cite{speth2021deception} proposed a multimodal deception database DDPM contains almost 13 hours of recordings of 70 subjects, as well as physiological signals such as thermal video frames and pulse oximeter data. Most studies on deception detection are designed and evaluated on private datasets, typically with relatively small sample sizes, and MDPE dataset addresses these drawbacks. Table 1 compares the sample size and length  for existing datasets and MDPE.

\textbf{Multimodal Deception Detection} \  Decades of research in psychology, and deception detection have documented verbal and nonverbal behavioral cues indicative of deceptive communication. Visual cues such as the frequency and duration of eye blinks \cite{bhaskaran2011lie,fukuda2001eye,minkov2012comparison}, dilation of pupils \cite{dionisio2001differentiation,lubow1996pupillary}, and facial muscle movements \cite{hurley2011executing,porter2011would} have been found to distinguish between deceptive and truthful behavior. Vocal cues can be indicative of deception, with deceptive speakers tending to speak with higher and more varied pitch \cite{depaulo2003cues,zuckerman1981verbal}, shorter utterances, and less fluency \cite{rockwell1997voice,sporer2006paraverbal} than truthful speakers. Deception also correlates with verbal attributes of speech, with deceivers tending to communicate with less cognitive complexity, fewer self-references, and more words indicative of negative emotions \cite{zhou2004automating,newman2003lying}. 
Mohamed et al. \cite{abouelenien2016detecting} explored a multimodal deception detection approach and integrates multiple physiological, linguistic, and thermal features. They used a decision tree model, to gain insights into the features that are most effective in detecting deceit.
Leena Mathur et al. \cite{mathur2020introducing} analyzed the discriminative power of features from visual, vocal, and verbal modalities affect for deception detection. They experimented with unimodal Support Vector Machines (SVM) and SVM-based multimodal fusion methods to identify effective features for detecting deception.

\textbf{Individual Difference Deception} \  Some studies confirm that some of the five NEO-FFI (Neuroticism-Extraversion-Openness Five-Factor Inventory) dimensions are related to deception \cite{ramanaiah1994revised,jakobwitz2006dark}. Sarah Ita Levitan et al. \cite{levitan2015cross} reported the role of personality factors derived from the NEO-FFI and of gender, ethnicity and confidence ratings on subjects’ ability to deceive and to detect deception. Justyna Sarzy{\'n}ska et al. \cite{sarzynska2017more} reports correlations between the ability to lie and extraversion, as well as conscientiousness. Personality characteristics are a promising set of information for deception detection, and similarly, emotional characteristics are also important. Joseph P. Gaspar et al. \cite{gaspar2022emotional} integrate prior theory and research on emotions, emotional intelligence, and deception and introduce a theoretical model. This model explores the interplay between emotional intelligence (the ability to perceive emotions, use emotions, understand emotions, and regulate emotions; and deception. 
Mircea Zloteanu et al hold strong beliefs about the role of emotional cues in detecting deception, and explored how decoders’ emotion recognition ability and senders’ emotions influence veracity judgements \cite{zloteanu2021veracity}.
Joseph P. Gaspar et al. \cite{gaspar2013emotion} believe that emotions are both an antecedent and a consequence of deception, and they introduce the emotion deception model to represent these relationships. This model broadens their understanding of deception in negotiations and accounts for the important role of emotions in the deception decision process.
To our knowledge, MDPE is the only deception detection dataset with personality and emotional characteristics.

\begin{figure*}[t]
\label{site}
      \centering
      \includegraphics[width=0.9\linewidth]{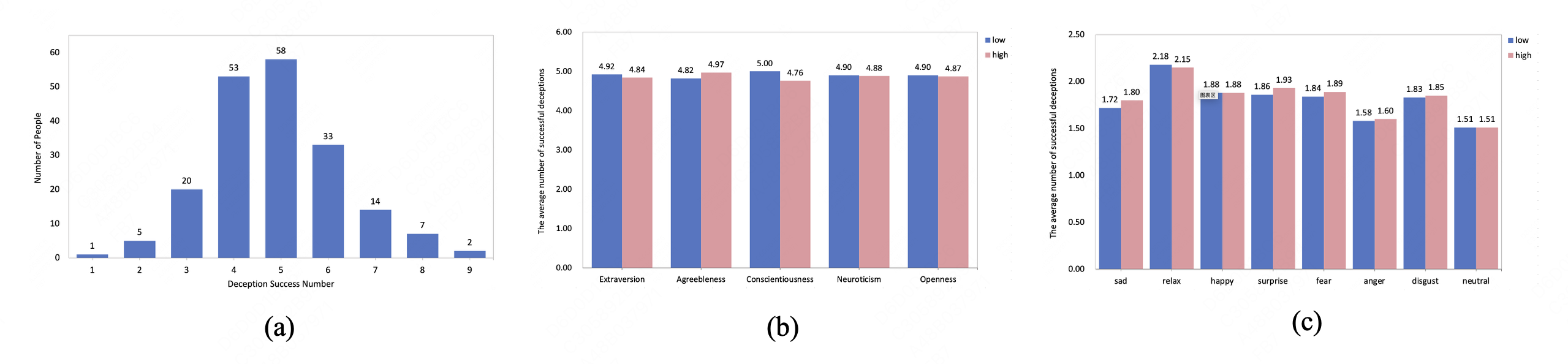}
      \caption{Statistical and analytical results of MDPE.}
  \end{figure*}

\section{Dataset}

\subsection{Experimental Setup}

\textbf{Equipment}: The dataset was collected using a GoPro Hero9 sports camera configured to record video at a resolution of 1920×1080 pixels and a frame rate of 60 frames per second (fps). Audio data were synchronously captured via the camera’s built-in microphone. During the emotional experiment, subjects were provided with a ThinkPad laptop to watch emotion-induction stimuli videos. Data collection took place in a controlled professional recording studio to minimize environmental interference. Only the participant and the interviewer remained in the room during the recordings. 

\textbf{Emotion-Induction Videos}: During the emotional induction experiment, subjects were shown a series of emotion-inducing videos designed to evoke specific emotional states, including sadness, happiness, relaxation, surprise, fear, disgust, anger, and neutrality. A total of 39 videos were utilized, with 17 collected from the Chinese Emotional Video System (CEVS) \cite{xu2010}. Each video segment in the CEVS has been professionally labeled and evaluated to ensure its effectiveness in eliciting the corresponding emotional responses. However, the CEVS only includes six emotions: sadness, happiness, fear, disgust, anger, and neutrality, excluding relaxation and surprise. Additionally, some of the CEVS videos failed to reliably evoke the intended emotional responses during our pre-experiment. This may be attributed to shifts in aesthetic preferences over time, resulting in reduced emotional resonance among contemporary viewers. To address this problem, an additional 22 videos were collected from online sources. These videos were annotated by 12 independent data annotators according to the same criteria and annotation methods as those used in the CEVS. The results showed that each video successfully elicited strong emotional responses.

\textbf{Deception Questions}: During the deception experiment, participants were asked a set of 24 "deception questions". These questions were developed by a panel of five psychology researchers, each with over five years of experience, drawing upon theoretical frameworks such as the Fraud Triangle Theory and Rational Choice Theory. To ensure comprehensiveness, the questions were designed to integrate interdisciplinary perspectives from psychology, criminology, and sociology, thereby capturing diverse dimensions of deceptive behavior. Some questions were specifically designed to reflect emerging trends in deceptive practices, addressing aspects potentially overlooked by conventional methodologies. The initial question set underwent a preliminary round of testing, during which participant feedback was collected and incorporated into subsequent revisions. The finalized version of the deception questions is provided in Appendix C.

\textbf{Personnel}: The study involved two distinct personnel roles: the interviewer and the Data Collection Coordinator (DCC). Interviewers were researchers with at least three years of experience in psychology and had received specialized training in deception detection techniques. Their responsibilities included conducting interviews, making real-time judgments regarding participants’ truthfulness, and completing the \textit{Interviewer Judgment Scale}. The data collection coordinator will assist with the execution of the study, including the preparation of materials and other related tasks. 

\textbf{Other Material}: The \textit{Interviewer Judgment Scale} assessed the interviewer's perception of each subject's response credibility, rated on a 5-point Likert scale (1 = definitely true, 5 = definitely false). After the interview, subjects completed the \textit{Confidence Scale} for Lying to self-assess their deceptive performance, also using a 5-point Likert scale (1 = I definitely deceived successfully, 5 = I definitely did not deceive successfully). During the emotional experiment, subjects rated their experience of eight specific emotions on an \textit{Emotional Scale}, using a 5-point scale (1 = no such emotion, 5 = strongest emotion). Details can be found in Appendix B.

\vspace{-0.15cm}
\subsection{Procedure}
Each subject was required to complete the following three experiments: personality, emotion and deception experiment.

\textbf{Personality Characteristics Collection}: Subjects were required to complete a Big Five personality questionnaire \cite{zhang2022big}, which consists of 60 items. Each item was rated on a Likert scale from 1 to 5, with 1 indicating strong disagreement and 5 indicating strong agreement, reflecting the degree to which each statement matched the subject's own characteristics. Based on the scoring methodology, the Big Five personality traits—openness, conscientiousness, extraversion, agreeableness, and neuroticism—were assessed for each subject. The full questionnaire and scoring methodology are provided in Appendix A.

\textbf{Emotional Experiment}: The DCC randomly selected 16 induction videos (ensure 2 videos for each emotion) for the subjects to watch. After watching each video, subjects were required to describe their feelings and then fill out an \textit{Emotional Scale}.

\textbf{Deception Experiment}: The deception data collection process follows DDPM \cite{speth2021deception}. The DCC randomly selected 9 questions that must lie and hand them over to the subject (the interviewer does not know which 9 questions). The first 3 questions will not be selected, which means that the first three “warm up” questions were always to be answered honestly. They allowed the subject to get settled, and gave the interviewer an idea of the subject’s demeanor when answering a question honestly. The subject have a maximum of 15 minutes to prepare, and during the preparation process, they must remember these 9 questions and think about how to deceive in the upcoming interview process. During the interview process, when asked these 9 questions, the subject must lie, and when asked the remaining 15 questions, they must tell the truth. Subjects were motivated to deceive successfully through two levels of bonus compensation: if they were able to deceive the interviewer in five or six of the nine deceptive responses, they were given a 150\% of a base incentive payment; the base payment was doubled if they were successfully deceptive in seven or more questions. In order to collect more indistinguishable deception answers, we encourage subjects to incorporate some truth into lies when answering these deceptive questions. 

During the interview process, the interviewer asked 24 questions in random order, and provide their judgment of truthful or deceptive answers to each question. And the interviewer filled out the \textit{Interviewer Judgment Scale}. And after the interview, the subject also filled out the \textit{Subject Lie Confidence Scale}.

\subsection{Statistic and Analysis}
A total of 193 subjects took part in this study, comprising 130 females and 63 males, all of whom were native Chinese speakers. Their ages ranged from 18 to 69 years, and their occupations included students, laborers, teachers, retirees, and others. Each subject contributed responses to 24 deception questions, yielding a total of 1737 deceptive and 2895 truthful responses.  In addition, each participant provided 16 emotion-inducing videos, resulting in a total of 3088 emotional video recordings. Following data collection, the raw video recordings were segmented. The duration of individual deceptive video clips ranged from 4 to 27 minutes, while emotional videos ranged from 19 to 38 minutes, which included the time spent watching emotion-induction materials. In total, 1808 minutes of deceptive video and 4401 minutes of emotional video were obtained, amounting to 6209 minutes of video content.

A preliminary analysis of the data reveals the distribution of successful deception attempts across all subjects (Figure 1 (a)). The majority of participants had success rates in the 3-6 times range. Figure 1 (b)  displays the average number of successful deception by personality category. Subjects were categorized into high and low groups for each Big Five personality trait based on mean scores. Analysis indicates no significant difference in deception success rates between high/low Neuroticism or Openness. However, significantly higher deception success rates were observed among individuals with low Extraversion, high Agreeableness, and low Conscientiousness. The finding regarding low Extraversion may be attributed to a tendency towards greater caution and reduced likelihood of revealing vulnerabilities through exaggeration. Individuals scoring high in Agreeableness, characterized by cooperativeness and trustfulness, may more easily gain trust and goodwill, potentially lowering interviewer vigilance. Similarly, those with low Conscientiousness typically place less emphasis on rules, obligations, and social norms. These patterns align with established perspectives from disciplines such as self-control theory and general crime theory \cite{collins1993personality, detert2008moral, gottfredson1990general}. Figure 1 (c) presents the average intensity of each emotion expressed during the emotion experiment, comparing subjects with high versus low deception success rates. Individuals exhibiting higher deception success rates demonstrated heightened emotional expressivity. This may reflect a greater capacity among successful deceivers to understand and simulate others' emotions, resulting in a more sensitive and reactive emotional system. This observation is consistent with existing theories linking emotion and cognition \cite{austin2007emotional, grandey2003show}.

\subsection{Ethics Review and License}

Before the experiment began, the subjects were informed of all experimental procedures. The subjects explicitly consented to record their conversation and publish the video data in a scientific conference or journal. And we do not publish any privacy-sensitive data, and the anonymity of participants will be guaranteed. All data were collected under a protocol approved by the authors’institution’s Human Subjects Institutional Review Board.

Additionally, we restrict the use of this dataset under the license of CC BY-NC-SA-4.0, requiring researchers to use our dataset responsibly. And commercial usage is prohibited.

\section{Benchmark}

\subsection{Data Preprocessing}

For the visual modality, raw videos are standardized to 30 fps. Faces are cropped and aligned using the DLib Toolkit [1]. Frame-level features are extracted using visual encoders and compressed to video-level via average pooling. For audio, the track is separated using FFmpeg, standardized to 16kHz mono, and acoustic features extracted. For text, transcripts are generated using the Paraformer ASR toolkit [2]. For each sample $x_i$, this yields acoustic features $f_{i}^{a} \in \mathbb{R}^{d_a}$, textual  features $f_{i}^{l} \in \mathbb{R}^{d_l}$, and visual features $f_{i}^{v} \in \mathbb{R}^{d_v}$, where $\left\{ d_m \right\} _{m\in\left\{ a,l,v \right\}}$ is the feature dimension for each modality.

\subsection{Feature Extraction}

Feature selection significantly impacts model performance. To guide feature selection, we evaluate distinct feature types under consistent experimental conditions. For visual modality, compared with handcrafted features, deep features extracted from supervised models are useful for facial expression recognition \cite{li2020deep}. CLIP \cite{radford2021learning} is a multimodal model based on contrastive learning, where training utilizes text and images to construct positive and negative sample pairs. The Vision Transformer (VIT) \cite{dosovitskiy2020image} is a transformer encoder model, pre-trained in a supervised manner on a large dataset of images. For acoustic modality, the Wav2vec2 \cite{baevski2020wav2vec} masks speech inputs in the latent space and addresses a contrastive task defined on quantized latent representations. It has been widely applied to downstream speech tasks. HUBERT \cite{hsu2021hubert} utilizes offline clustering steps to provide aligned target labels for prediction losses. To better distinguish between speakers, a sentence mixture training strategy WavLM \cite{chen2022wavlm}  is proposed, allowing for the unsupervised creation and merging of additional overlapping sentences during the training process. For text modality, we extract the sbert-chinese-general-v2 features, which is based on the bert-base-chinese version of the BERT model. ChatGLM \cite{du2021glm} is an open-source conversational language model, based on the General Language Model (GLM) architecture. It demonstrats exceptional contextual understanding and more efficient inference capabilities. Baichuan \cite{yang2023baichuan} is an open-source large-scale model with 13 billion parameters. It features a larger size, more extensive training data, and more efficient inference capabilities.



\subsection{Model Structure}

For unimodal features, we utilize the fully-connected layers to extract hidden representations and predict deception:

 \begin{equation}
h_{i}^{m}=\text{ReLU}\left( f_{i}^{m}W_{m}^{h}+b_{m}^{h} \right) ,{m\in \left\{ a,l,v \right\}}
 \end{equation}
   \begin{equation}
\hat{y}_i = \text{softmax} \left( h_{i}^{m}W_{m}^{d}+b_{m}^{d} \right) ,m\in \left\{ a,l,v \right\} 
 \end{equation}

where $h_{i}^{m} \in \mathbb{R}^h $ is the hidden feature for each modality, $d_{i} \in \mathbb{R}^2 $ is the estimated deception probabilities. For multimodal features, different modalities contribute differently to deception detection. Therefore, we compute importance scores $\alpha _i \in \mathbb{R}^{3\times1}$ for each modality and exploit weighted fusion to obtain multimodal features:

   \begin{equation}
h_i= \text{Concat}\left( h_{i}^{a},h_{i}^{l},h_{i}^{v} \right) 
 \end{equation}
   \begin{equation}
\alpha _i=\text{softmax} \left( h_{i}^{T}W_{\alpha}+b_{\alpha} \right) 
 \end{equation}

Similarly, for personality traits and emotional expression features, feature fusion is achieved through concatenation. Personality features are derived directly from personality scale scores. Emotional expression features, however, are obtained by first training a dedicated emotion recognition model. Subsequently, all emotional expression samples are processed through this model, and the activations from the last fully connected layer are extracted. These activations then undergo average pooling to form the final emotion expression feature vector. It is noted that the model architecture employed for both personality recognition and emotion recognition tasks remains identical.

\begin{table}[b]
\caption{Unimodal results of deception detection. "P" denotes the addition of personality features and "E" denotes the addition of emotional features.}
\begin{tabular}{ccccc}
\hline
Feature        & Accuracy         & with P           & with E           & with P \& E     \\ \hline
VIT            & 60.30\%          & 61.27\%          & 61.43\%          & 61.55\%          \\
CLIP-base      & 58.54\%          & 59.17\%          & 58.32\%          & 59.11\%          \\
CLIP-large     & 57.30\%          & 58.34\%          & 56.97\%          & 57.67\%          \\ \hline
eGeMAPS        & 55.86\%          & 57.22\%          & 56.22\%          & 56.89\%          \\
HUBERT-base    & 58.13\%          & 62.38\%          & 59.35\%          & 62.12\%          \\
HUBERT-large   & 60.80\%          & 62.07\%          & 60.34\%          & 61.87\%          \\
Wav2vec2-base  & 58.75\%          & 59.74\%          & 59.99\%          & 59.84\%          \\
Wav2vec2-large & 60.10\%          & 61.88\%          & 59.32\%          & 62.10\%          \\
WavLM-base     & 61.66\%          & 60.82\%          & 60.16\%          & 60.92\%          \\
WavLM-large    & 57.82\%          & 60.31\%          & 58.02\%          & 60.52\%          \\ \hline
Sentence-BERT  & 61.76\%          & 62.34\%          & 63.21\%          & 63.34\%          \\
ChatGLM2-6B    & 60.73\%          & 61.45\%          & 61.45\%          & 61.56\%          \\
Baichuan-13B   & \textbf{61.87\%} & \textbf{62.90\%} & \textbf{63.32\%} & \textbf{63.74\%} \\ \hline
\end{tabular}
\end{table}

\subsection{Implementation Details}

The dimension of latent representations is selected from $\left\{64, 128, 256\right\}$. We employ the Adam optimizer \cite{kingma2014adam} with a learning rate chosen from $\left\{10^{-3}, 10^{-4}\right\}$, a weight decay of $10^{-5}$, and a maximum of 300 training epochs. To mitigate overfitting, dropout \cite{srivastava2014dropout} is applied, with rates selected from $\left\{0.2, 0.3, 0.4, 0.5\right\}$. The cross-entropy loss function is used for optimization. For deception detection tasks, each sample comprises 24 answers. We randomly select 5 answers per sample (3 truthful, 2 deceptive) for validation, reserving the remaining 19 for training. All experiments are repeated five times with randomized initializations, and results report average performance to ensure statistical reliability. For personality and emotion recognition tasks, the dataset is split into 133 training samples, 40 validation samples, and 40 test samples; these tasks utilize root mean square error (RMSE) as the loss function.

For deception detection, accuracy was selected as the evaluation metric. For personality recognition, we employed the mean accuracy (A), defined as follows:
 
 \begin{equation}
 A = 1 - \frac{1}{N^t} \sum_{i}^{N^t} \left| \mathbf{Y}_i^P - \mathbf{P}_i \right| 
 \end{equation}

This metric is widely adopted in personality recognition tasks \cite{ponce2016chalearn, zhang2019persemon}. For emotion recognition, the root mean square error (RMSE) was used.

\begin{table}[t]
\caption{Multimodal results of deception detection.}
\begin{tabular}{ccccccc}
\hline
V   & A   & T   & Accuracy         & with P           & with E           & with P \& E     \\ \hline
VIT & HBB & -   & 61.76\%          & 61.53\%          & 61.14\%          & 62.68\%          \\
VIT & WMB & -   & 60.88\%          & 60.22\%          & 60.02\%          & 60.72\%          \\
CLB & HBB & -   & 61.02\%          & 60.14\%          & 60.53\%          & 61.45\%          \\ \hline
VIT & -   & Bai & 63.31\%          & 63.47\%          & 63.31\%          & 63.48\%          \\
CLB & -   & Bai & 63.31\%          & 63.52\%          & 63.38\%          & 63.10\%          \\
CLL & -   & Bai & 64.15\%          & 63.94\%          & 63.98\%          & 64.04\%          \\ \hline
-   & HBL & Bai & 62.69\%          & 63.42\%          & 63.00\%          & 63.42\%          \\
-   & W2B & Bai & 63.83\%          & 63.48\%          & 63.57\%          & 63.79\%          \\
-   & WMB & Bai & 64.25\%          & 64.15\%          & 63.90\%          & 64.07\%          \\ \hline
VIT & HBB & Bai & \textbf{64.45\%} & 64.33\%          & 63.93\%          & 64.00\%          \\
VIT & WMB & Bai & 63.42\%          & \textbf{64.87\%} & 63.62\%          & 63.59\%          \\
CLB & HBB & Bai & 62.94\%          & 63.93\%          & \textbf{63.97\%} & \textbf{64.66\%} \\ \hline
\end{tabular}
\end{table}

\subsection{Experiment Results}

This section establishes the benchmark for MDPE, designed to guide feature selection and inform the development of robust feature extractors.

Unimodal deception detection results (Table 2) indicate that the textual modality significantly outperforms visual and acoustic modalities. This suggests that textual cues in our dataset provide more salient deceptive indicators. Subsequent multimodal fusion experiments (Table 3) reveal that integrating complementary modalities consistently enhances performance. This improvement aligns with the premise that deception manifests across multiple channels, enabling models to better comprehend video content and detect deception more accurately. Notably, while most unimodal features benefit from fusion, combining visual and acoustic features yields negligible gains or performance degradation. This implies that textual features stabilize model performance, a finding consistent with human deception judgment, where content-based (textual) cues typically dominate over visual or acoustic signals. Further exploration of deceptive indicators within visual and acoustic modalities is warranted.

Incorporating personality features consistently improves deception detection performance, underscoring their importance for the task. While emotion features also enhance performance, their contribution is less pronounced than personality features and occasionally detrimental. This discrepancy may arise because personality traits serve as direct indicators, whereas emotion features depend on the quality of upstream emotion recognition models. Future work should explore more effective methods for leveraging emotional expression features. Combining both personality and emotion features achieves the highest unimodal deception detection performance, confirming their relevance and demonstrating the viability of individual-difference-based modeling.

For personality recognition (Table 3), acoustic features surpass visual/textual features in isolation, highlighting the predictive value of vocal cues. Full multimodal fusion achieves optimal performance, indicating cross-modal complementarity. In emotion recognition, textual features provide the strongest cues, and multimodal integration significantly outperforms unimodal approaches, emphasizing the necessity of cross-modal fusion for robust emotion understanding.

\begin{table}[]
\caption{Experiment results of personality recognition and emotion recognition.}
\begin{tabular}{cllcccc}

\hline
\multicolumn{3}{c}{\multirow{2}{*}{Feature}}                            & \multicolumn{2}{c}{Personality} & \multicolumn{2}{c}{Emotion}                \\ \cline{4-7} 
\multicolumn{3}{c}{}                                                    & Val            & Test           & Val            & Test                      \\ \hline
\multicolumn{3}{c}{VIT}                                                 & 90.70\%         & 92.2\%           & 1.288          & 1.351                     \\
\multicolumn{3}{c}{ClipVIT-B16}                                         & 92.06\%         & 91.75\%          & \textbf{1.272}          & 1.333                    \\
\multicolumn{3}{c}{ClipVIT-L14}                                         & 91.81\%         & 91.47\%           & 1.277          & 1.339                     \\ \hline
\multicolumn{3}{c}{HUBERT-base}                                         & 92.01\%         & 92.65\%          & 1.286         & \textbf{1.330}                     \\
\multicolumn{3}{c}{HUBERT-large}                                        & 91.62\%         & 92.66\%           & 1.290          & 1.336                     \\
\multicolumn{3}{c}{Wav2vec2-base}                                       & 92.27\% & \textbf{93.16\%}          & 1.281 & 1.337                     \\
\multicolumn{3}{c}{Wav2vec2-large}                                      & 92.32\%          & 92.96\%          & 1.287          & 1.342                     \\
\multicolumn{3}{c}{WavLM-base}                                          & 91.64\%         & 92.53\%            & 1.280
         & 1.336                    \\
\multicolumn{3}{c}{WavLM-large}                                         & 91.82\%          & 92.55\%         & 1.287          & 1.344                     \\ \hline
\multicolumn{3}{c}{Sentence-BERT}                                       & 92.23\%         & 93.04\%          & 1.274         & 1.327                    \\
\multicolumn{3}{c}{ChatGLM2-6B}                                         & 91.75\%         & 92.63\%           & 1.276          & 1.334                     \\
\multicolumn{3}{c}{Baichuan-13B}                                        & \textbf{92.30\%}         & 93.11\% & 1.274         & 1.332 \\ \hline
\multicolumn{1}{l}{VIT} & W2V                   & \multicolumn{1}{c}{-} & 90.71\%         & 92.36\%         & 1.272          & 1.334                     \\
-                       & W2V                   & BAI                   & \textbf{92.34}\%         & 93.35\%           & 1.273          & 1.332                    \\
\multicolumn{1}{l}{VIT} & \multicolumn{1}{c}{-} & BAI                   & 90.74\%         & 92.31\%   & 1.275          & 1.334 \\
\multicolumn{1}{l}{VIT} & W2V                   & BAI                   & 92.16\%         & \textbf{93.43}\%            & \textbf{1.270}          & \textbf{1.229}                     \\ \hline
\end{tabular}
\end{table}

\section{Limitations}

Firstly, although the subjects were required that they must lie about the deception  questions, and verified the deceptive questions and content with the Interviewer after the deception experiment, we do not know whether the subjects have actually deceived on the deception  questions. Secondly, relying on self-assessment scales for data annotation is a subjective process for subjects, which may lead to bias in subsequent analysis. Different subjects may have significant differences in their perception of emotions. In addition, MDPE only collects native Chinese speakers, there may be cultural differences in deception detection. Finally, gender imbalance among subjects in MPDE is a common issue in human data collection \cite{d2022exclusion,pinho2022dementia}.

\section{Conclusion}

We introduce the Multimodal Deception Detection Dataset (MDPE), comprising video, audio, and textual modalities, supplemented with personality and emotion annotations. This dataset enables cross-modal analysis to investigate complementary relationships between modalities, thereby advancing robust deception detection methodologies relevant to societal security. Furthermore, MDPE facilitates research into the influence of personality traits and emotional states on deceptive behavior. Beyond its primary application, the dataset supports auxiliary tasks such as personality and emotion recognition, as well as joint analyses of deception-personality-emotion interactions. Benchmark experiments are provided to ensure reproducibility and establish a foundation for future work. By publicly releasing MDPE, we aim to stimulate progress in this critical area of affective computing.

\bibliographystyle{ACM-Reference-Format}

\newpage
\appendix
\onecolumn

\section{Big Five Personality Inventory Second Edition (BFI-2)}

Below are some descriptions of personal characteristics, some may or may not apply to you. Please fill in the corresponding number on the horizontal line before each sentence below to indicate whether you agree or disagree with this description.

\begin{enumerate}

 \item Outgoing personality, enjoys socializing
 \item Soft hearted and compassionate
 \item Lack of organization
 \item Calm and adept at handling pressure
 \item Not very interested in art
 \item Strong and confident personality, daring to express one's own opinions
 \item Humble and respectful towards others
 \item Relatively lazy
 \item Being able to maintain a positive attitude even after experiencing setbacks
 \item Interested in many different things
 \item I rarely feel excited or particularly want to do anything
 \item Often picking on others' faults
 \item Reliable and reliable
 \item Irregular mood and frequent emotional fluctuations
 \item Skilled in creativity and able to find smart ways to do things
 \item Relatively quiet
 \item Lack of empathy towards others
 \item Work in a planned and organized manner
 \item Easy to get nervous
 \item Enthusiastic with art, music, or literature
 \item Often in a dominant position, like a leader
 \item Often having disagreements with others
 \item It's difficult to start taking action to complete a task
 \item Feeling secure and satisfied with oneself
 \item Disliking discussions with strong knowledge or philosophy
 \item Not as energetic as others
 \item Be magnanimous and magnanimous
 \item Sometimes I lack a sense of responsibility
 \item Emotionally stable and less likely to get angry
 \item Almost no creativity
 \item Sometimes shy and introverted
 \item Helpful and selfless towards others
 \item Habit keeps things tidy and orderly
 \item Often worried and worried about many things
 \item Valuing Art and Aesthetics
 \item Feeling difficult to influence others
 \item Sometimes being rude to people
 \item Efficiency, starting and ending with work
 \item Often feeling sad
 \item Deep thinking
 \item Full of energy
 \item Do not trust others and doubt their intentions
 \item Reliable, always trustworthy to others
 \item Able to control one's emotions
 \item Lack of imagination
 \item Loud and talkative
 \item Sometimes cold and indifferent to others
 \item It's messy and doesn't like to tidy up
 \item Rarely feel anxious or afraid
 \item Feeling bored with poetry and drama
 \item I prefer to have others take the lead and take responsibility
 \item Humility and courtesy towards others
 \item Have perseverance and be able to persist in completing tasks
 \item Often feeling depressed and unhappy
 \item Not very interested in abstract concepts and ideas
 \item Full of enthusiasm
 \item Think about people in the best possible way
 \item Sometimes they may engage in irresponsible behavior
 \item Emotions are variable and prone to anger
 \item Creative and able to come up with new ideas

\end{enumerate}

\section{Emotional Sacle}

You need to rate the following emotions: sadness, relaxation, happiness, surprise, fear, anger, disgust, and neutral. Mark to what extent you feel it appropriately expresses your feelings, with intensity ranging from 1 to 5, where 1 is the least intense and 5 is the strongest.

\begin{table}[h]

\begin{tabular}{|c|c|c|c|c|c|c|c|c|}
\hline
Video Number & Sadness & Relax & Happiness & Surprise & Fear & Angry & Disgust & Neutral \\ \hline
1            &         &       &           &          &      &       &         &         \\ \hline
2            &         &       &           &          &      &       &         &         \\ \hline
3            &         &       &           &          &      &       &         &         \\ \hline
4            &         &       &           &          &      &       &         &         \\ \hline
5            &         &       &           &          &      &       &         &         \\ \hline
6            &         &       &           &          &      &       &         &         \\ \hline
7            &         &       &           &          &      &       &         &         \\ \hline
8            &         &       &           &          &      &       &         &         \\ \hline
9            &         &       &           &          &      &       &         &         \\ \hline
10           &         &       &           &          &      &       &         &         \\ \hline
11           &         &       &           &          &      &       &         &         \\ \hline
12           &         &       &           &          &      &       &         &         \\ \hline
13           &         &       &           &          &      &       &         &         \\ \hline
14           &         &       &           &          &      &       &         &         \\ \hline
15           &         &       &           &          &      &       &         &         \\ \hline
16           &         &       &           &          &      &       &         &         \\ \hline
\end{tabular}
\end{table}

\section{Interview Questions}

\begin{enumerate}
\item What color do you like the most? Why?
\item Where is your hometown? Please briefly introduce it.
\item Do you have any hobbies?
\item Have you traveled in the past year?
\item How do you like Beijing?
\item What is your happiest experience?
\item What is your favorite food?
\item What is your personality like?
\item What is your biggest weakness?
\item What is your greatest strength? 
\item What do you usually do to relax?
\item Which exercise or sport do you like?
\item Briefly introduce your family members.
\item Who is the person you have the greatest influence on you?
\item Do you have any special places or tourist destinations you want to go to?
\item Who is your favorite celebrity or great person?
\item What is your opinion on the words "neijuan" and "tangping"?
\item What is your favorite literary and artistic work?
\item Have you ever received any rewards or honors in school or at work?
\item What was your most unforgettable experience in the past year?
\item Have you participated in any major event?
\item Have you ever cheated in school or work?
\item Have you concealed a fact to your family or friends in the past year?
\item Have you ever lied to avoid responsibility?
\end{enumerate}

\end{document}